%% file: main.tex
\pdfoutput=1

\documentclass[11pt]{article}

\usepackage[]{acl}

\usepackage{times}
\usepackage{latexsym}
\usepackage{amsmath}
\usepackage{algorithm} 
\usepackage{algpseudocode} 
\usepackage[T1]{fontenc}
\usepackage{fvextra} 
\usepackage{tabularx}
\usepackage{amssymb}
\usepackage[utf8]{inputenc}
\usepackage{color}
\usepackage{microtype}
\usepackage{graphicx}
\usepackage{algpseudocode}
\usepackage{xcolor}
\usepackage{float}

\usepackage{listings}

\usepackage{hyperref} 
\title{Lossless Acceleration of Large Language Model via Adaptive N-gram Parallel Decoding}

\author{Jie Ou, Yueming Chen,  Wenhong Tian\thanks{ ~~Corresponding author} \\
  University of Electronic Science and Technology of China, Chengdu, China \\
  \texttt{oujieww6@gmail.com, yuemingchen121@gmail.com}\\
  \texttt{tian\_wenhong@uestc.edu.cn}}

\begin{document}
\maketitle
\begin{abstract}
While Large Language Models (LLMs) have shown remarkable abilities, they are hindered by significant resource consumption and considerable latency due to autoregressive processing. In this study, we introduce Adaptive N-gram Parallel Decoding (ANPD), an innovative and lossless approach that accelerates inference by allowing the simultaneous generation of multiple tokens. ANPD incorporates a two-stage approach: it begins with a rapid drafting phase that employs an N-gram module, which adapts based on the current interactive context, followed by a verification phase, during which the original LLM assesses and confirms the proposed tokens. Consequently, ANPD preserves the integrity of the LLM's original output while enhancing processing speed. We further leverage a multi-level architecture for the N-gram module to enhance the precision of the initial draft, consequently reducing inference latency. ANPD eliminates the need for retraining or extra GPU memory, making it an efficient and plug-and-play enhancement. In our experiments, models such as LLaMA and its fine-tuned variants have shown speed improvements up to \(3.67\times\), validating the effectiveness of our proposed ANPD.

\end{abstract}

\input{intro}
\input{related}
\input{method}
\input{exps}
\input{conc}

\bibliography{anthology,custom}
\bibliographystyle{acl_natbib}

\end{document}

%% file: intro.tex
\section{Introduction}
The advent of Large Language Models (LLMs) such as GPT-4 \citep{openai2023gpt4}, ChatGPT \citep{brown2020language}, LLaMA \citep{touvron2023llama}, and PaLM \citep{chowdhery2022palm}, has revolutionized the landscape of natural language processing. However, the majority of LLMs \cite{touvron2023llama, anil2023palm, bai2023qwen} rely on the decoder-only Transformers architecture \cite{alec2018improving}, which is intrinsically autoregressive and consequently leads to increased generation time during inference. This characteristic has made the improvement of LLM inference efficiency a significant research area within the natural language processing community.

Model compression techniques such as quantization \citep{han2015deep}, pruning \citep{molchanov2016pruning}, and distillation \citep{hinton2015distilling} have been employed to alleviate the computational costs associated with LLMs.  Recently, innovative methods such as early exit strategies \citep{yang2023predictive, bae2023fast, kong2022accelerating, schuster2022confident, varshney2023accelerating} and speculative decoding \citep{kim2023speculative,xia2022speculative,leviathan2023fast,spector2023accelerating,zhang2023draft} have been proposed to speed up the inference process. While these methods are effective, they typically necessitate modifications to the model architecture and re-training, which can incur substantial costs. Additionally, they may alter the model's output and require extra GPU memory needs. A method avoiding draft models using retrieval is presented in \citep{he2023rest}, but it requires a large database.

\begin{figure}[h!]
    \centering
    \includegraphics[width=0.4\textwidth]{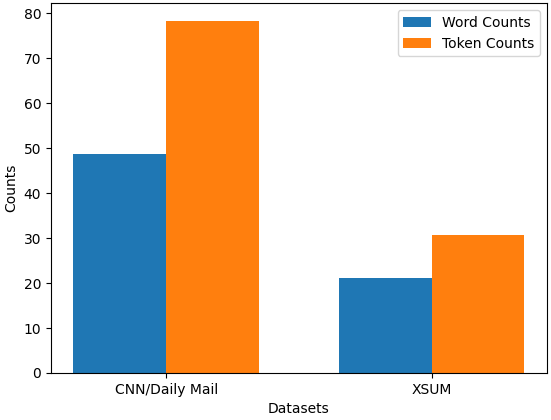}
    \caption{The comparative analysis of the number of words and tokens after tokenizer processing for the CNN/Daily Mail and XSUM datasets.}
    \label{fig:1}
\end{figure}

For certain LLMs, such as LLaMA, the tokenization process can dissect a single word into multiple tokens, thereby exacerbating inference latency. As illustrated in Figure \ref{fig:1}, the token count exceeds the word count, resulting in an increased number of autoregressive generation steps. In such scenarios, given the constraints imposed by contextual information, the search space for predicting the next token that forms part of a word based on the current token is significantly narrowed. Moreover, contextual information can often be leveraged to identify patterns and correlations between words. This is especially evident for simple phrases and paragraphs, where the context can provide clear indicators that reduce the dependency on LLM decoding.

Based on the above motivation,  this paper presents a novel approach,  the \textbf{A}daptive \textbf{N}-gram \textbf{P}arallel \textbf{D}ecoding (\textbf{ANPD}), designed to enhance inference efficiency without necessitating retraining or the integration of an auxiliary small language model. ANPD dynamically generates draft outputs via an adaptive N-gram module using real-time statistics, after which the drafts are verified by the LLM. This characteristic is exactly the difference between ANPD and the previous speculative decoding methods. The primary contributions of this work can be summarized as follows:

\begin{itemize}
    \item We propose ANPD, a novel and lossless algorithm that offers a plug-and-play module for acceleration of LLM inference. 
    \item We propose an adaptive N-gram modeling strategy that is specifically adapted for LLMs, markedly diminishing the complexity of language modeling and reducing the dependency on large-scale textual datasets.
    \item We propose a Multi-Level N-gram (MLN) algorithm aimed at increasing the precision of draft outputs, thereby enhancing the efficiency of the acceleration process.
    \item We conduct extensive experiments on various models and datasets, demonstrating the robust acceleration capabilities of ANPD, with a notable increase of 1.95$\times$-3.67$\times$ on LLaMA and its fine-tuned derivatives.
\end{itemize}

%% file: related.tex
\section{Related Work}

\textbf{Inference systems.} 
The development of specialized inference systems for Large Language Models (LLMs), such as NVIDIA's TensorRT-LLM \citep{NVIDIATensorRTLLM}, Orca \citep{yu2022orca}, FlexGen \citep{sheng2023high}, and DeepSpeed Inference \citep{aminabadi2022deepspeed}, represents a notable advancement in the field. Despite progress, there is still a gap in the careful co-design of algorithms and systems, which is necessary to fully harness the potential of the hardware.

\textbf{Compression.}
Efficient LLM inference is facilitated by techniques such as quantization \citep{han2015deep, frantar2022gptq, dettmers2022llm, xiao2023smoothquant}, pruning \citep{bansal2022rethinking, frantar2023massive, liu2023deja}, distillation \citep{tang2019distilling, touvron2021training}, and exit early strategies \citep{schuster2022confident, kong2022accelerating, yang2023predictive, bae2023fast, del2023skipdecode} suggest that some tokens can be accurately generated using only a fraction of the model layers. Token Prunings \citep{hou2022token, yao2022random, zhang2023h} reduce memory and computational demand to accelerate the inference process by prioritizing crucial tokens. These methods enhance efficiency but may necessitate model alterations, re-training, and potentially reduce accuracy.

\textbf{Speculative Execution.}
Speculative execution \citep{burton1985speculative}, adapted as speculative decoding in LLMs \citep{chen2023accelerating, leviathan2023fast}, has improved inference speeds by preempting computations. SpecInfer \citep{miao2023specinfer} leverages existing distilled, quantized, and pruned variants of an LLM, to build a small speculative model pool to guide speculation. However, these approaches require a high-quality draft model, and increase the memory footprint. \citet{leviathan2023fast} also mentioned that unigram and bigram can be used as draft models, but they did not propose a method on how to build a bigram model for the actual running LLMs. \citet{yang2023inference} presented a method of copying reference tokens to the decoder, though its utility is limited by a dependency on repeated text. These techniques increase resource use and compel specialized training, such as distillation, for the draft model to ensure compatibility with the primary model.

%% file: method.tex
\begin{figure*}[h!]
    \centering
    \includegraphics[width=\textwidth]{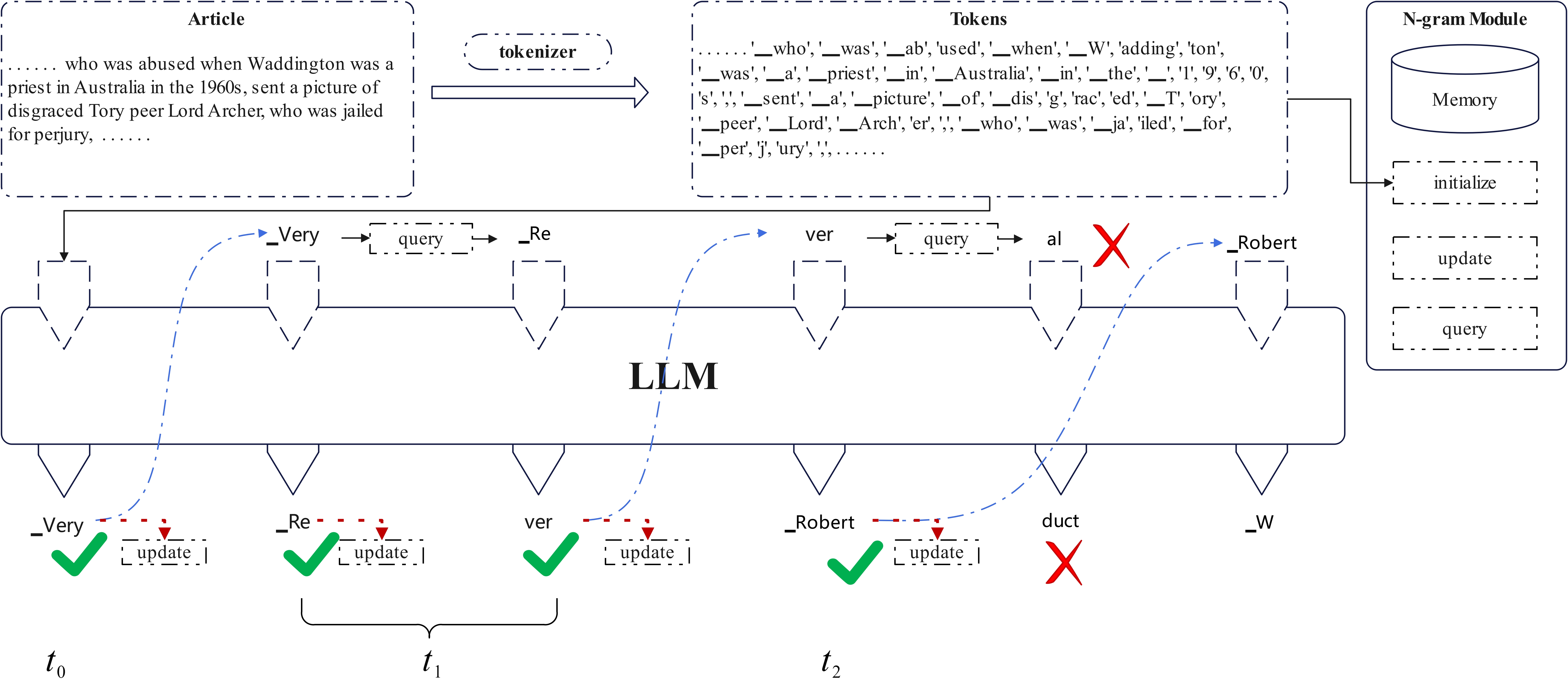}
    \caption{The pipeline of the ANPD. The tokenizer first processes the text to obtain a list of tokens. These tokens are used to initialize the N-gram module. Simultaneously, these tokens are fed into the LLM for processing via autoregression. The predicted token at time $t_0$ in the figure is "\_Very". This word is used as a query into the N-gram module, yielding the token "\_Re", which along with the "\_Very" are sent to the LLM for inference at time $t_1$. A \textcolor{green}{green checkmark}  signifies acceptance of the predicted token, while a \textcolor{red}{red cross} indicates rejection. Each accepted token, is combined with the first $N-1$ tokens to form a tuple, and the \textbf{update} method is called to refresh the N-gram module.}
    \label{fig-pipeline}
\end{figure*}
\section{Method}
Figure \ref{fig-pipeline}  illustrates the framework and workflow of proposed ANPD. We explain the original autoregressive decoding in the Appendix A.1.

\subsection{Adaptive N-gram Parallel Decoding}
Figure~\ref{fig-pipeline} illustrates the pipeline of our ANPD. The process begins with tokenizing the input text into tokens. The N-gram module's Memory actually stores token ids to streamline processing, Figure~\ref{fig-pipeline} shows tokens as the basis for modeling to make it easier for readers to understand and improve readability. Next, the LLM engages in autoregressive inference, divided into two parts: 1. Prefill, where the full prompt is input to generate the first token; 
2. Decoding, ANPD feeds multiple tokens from the N-gram module into the LLM, and the LLM uses kv-cache for efficient computations to validate tokens for parallel output generation. Tokens that fail validation are discarded along with subsequent tokens. Simultaneously, we use an adaptive strategy to update the N-gram module throughout LLM generation, avoiding reliance on static Memory.

\textbf{Token Level N-gram Module.} 
Contextual information is vital for content extraction, summarization, and code generation, as it helps refine the search space during each LLM decoding step. This includes strong correlations among tokens within words and between words in phrases and contexts. We constructed a token-level N-gram module to uniformly model the above correlations.
The N-gram module\footnote{https://web.stanford.edu/\textasciitilde jurafsky/slp3/3.pdf} is a probabilistic language model, that predicts the next item in a sequence using an $(N-1)$-th order Markov model, where $N$ is the subsequence length.
For a token sequence $x_1, x_2, ..., x_{t-1}$, the model estimates the probability of $x_{t}$ based on the preceding $N-1$ tokens, as $P(x_{t} | x_{1}, ..., x_{t-1}) \approx P(x_{t} | x_{t-N+1}, ..., x_{t-1})$. 
In a bigram model ($N=2$), the sentence probability is:
\begin{equation}
\label{eq2}
P(x_1, x_2, ..., x_n) \approx \prod_{i=2}^{n} P(x_i | x_{i-1}),
\end{equation}
probabilities $P(x_i | x_{i-1})$ derive from frequency counts in the corpus.
We have architected the N-gram module to encapsulate three principal functions essential for its operation: 
\begin{itemize}
    \item \textbf{Initialize:}  using a tokenizer converts each prompt into a sequence of token ids. It then performs probabilistic statistics on these ids and records the probability for each token tuple.
    \item \textbf{Update:} during the decoding, each new token is paired with the previous $N-1$ tokens to form a tuple, used to update the module's probability Memory.  
    \item \textbf{Query:} the query operation utilizes the token ids tuple, constructed through the subsequence from $t-N+1$ to $t-1$, to predict the next token $x_{t}$, effectively leveraging the statistical results established by the preceding functions. 
\end{itemize}
These functions collectively enable the N-gram module to dynamically adapt to the evolving text generation process, ensuring that each token generated is contextually relevant and statistically coherent.

\textbf{Parallel Decoding.}
The parallel decoding in our ANPD is similar to the speculative decoding approach and occurs in two distinct stages:
\begin{itemize}
    \item [1.] Drafting: the N-gram module is harnessed to generate a sequence of subsequent tokens. By iterating through $K$ steps, the module constructs a preliminary draft tokens with length $K$. Specifically, the draft module generates a series of $K$ temporary tokens $x_{i+1}, ..., x_{i+K}$, succeeding a given prompt sequence $x_1, ..., x_i$.
    \item [2.] Verification: the original Large Language Model (LLM) verifies the proposed draft tokens, through a singular forward pass as $P(x_{i+K+1}^{'}|(k,v)_1,...,(k,v)_i, x_{i+1}, ..., x_{i+K})$, within which the LLM computes the probability distributions for each draft token, then to ascertain their congruence with the proposed draft tokens $x_{i+1}, ..., x_{i+K}$. If a draft token $x_j$ does not pass this validation, it is replaced by the LLM's prediction $x_{j}^{'}$, and a new drafting begins from this token. 
\end{itemize}

The ANPD enhances efficiency by eliminating the need for a smaller draft deep learning model, leveraging the much lower computational cost N-gram module to accelerate LLM inference.
For LLMs, conducting parallel inference of $K$ tokens introduces a negligible increase in computational latency compared to single token autoregressive inference, as shown in Figure 7 in Appendix A.2. Meanwhile, 
our technique is intrinsically capable of yielding at least $j$ tokens ($1 \leq j \leq K+1$) for each decoding step, this intrinsic capability fundamentally assures, in principle, an acceleration of the decoding processes within the Large Language Model (LLM), thereby enhancing the overall computational throughput and reducing latency.
The implementation of the two-stage process confers upon the ANPD the ability to iteratively refine draft outputs. 
Furthermore, this guarantees that our ANPD method is lossless, maintaining consistency with the original LLM's generated content.
The detailed procedure of ANPD is presented in Algorithm 1, with a comprehensive explanation available in Appendix \ref{app:alg1_details}.
\begin{algorithm}[h!]
\caption{Adaptive N-gram Parallel Decoding}
\begin{algorithmic}[1]
\State \textbf{Input:}  $prompt$, $K$, $M$
\State \textbf{Output:} $O$
\State $token\_ids \gets \Call{tokenizer}{prompt}$
\State $Memory \gets \Call{Initialize}{token\_ids}$
\State $O \gets [~]$, $drafts \gets [~]$
\State $pred \gets \Call{LLM}{prompt}$ 
\State $drafts.append(pred[-1])$
\While{$length(O)$ \textless $M$}
    \State $token\_ids.append(drafts[1])$
    \State $O.append(token\_ids[-1]),\Call{Update}{O[-1]}$

    \State $tmp\_token\_ids \gets token\_ids[-N+1:]$
    \For{$k \gets 1$ to $K$}
        \State $tmp \gets tmp\_token\_ids[-N+k:]$ 
        \State $drafts.append(\Call{Query}{tmp})$ 
        \State $tmp\_token\_ids.append(drafts[-1])$
    \EndFor
    \State $predicts \gets \Call{LLM}{drafts}$ 
    \For{$j \gets 2$ to $\Call{length}{drafts}$}
        \If{$drafts[j] == predicts[j-1]$}
            \State $O.append(drafts[j])$
            \State $\Call{Update}{drafts[j]}$
            \State $token\_ids.append(drafts[j])$
        \Else
            \State \textbf{break}
        \EndIf
        
    
    \EndFor
    \If{$j == \Call{length}{drafts} $} 
        \State $drafts \gets [predicts[j]]$
    \Else
        \State $drafts \gets [predicts[j-1]]$
    \EndIf
\EndWhile
\end{algorithmic}
\end{algorithm}

\subsection{Multi-Level N-gram}
The predictive accuracy of the N-gram module is known to correlate with 
\( N \), larger \( N \) values generally result in more accurate content predictions. This effect is especially noticeable in settings with the longer context of Language Model (LM) tasks, where increasing \( N \) can markedly decrease the frequency of prediction errors.

While a larger \( N \)  tends to improve the predictive accuracy of the N-gram module, it may not always result in a successful match during the Query operation.  To address this, we propose the Multi-Level N-gram (MLN) approach, which is based on optimal prefix matching. The MLN design initializes \( N-1 \) separate modules, each corresponding to an $n$-gram module ($n\in [2, N]$). During prediction, the query starts with the largest \( N \) and proceeds to lower  \( n \)  levels, stopping when a successful match is found as shown in Algorithm 2.

\begin{algorithm}[t]
\caption{Multi-Level N-gram}
\begin{algorithmic}[1]

\State \textbf{Input:}  $tmp$, $N$,$token\_ids$ 
\State \textbf{Output:} $result$ 
\State $Memory \gets \Call{Initialize}{token\_ids}$ 

\State $result \gets \text{NULL}$ 
\State $n \gets N$ 

\While{$n \geq 2$}
    \State $pred \gets \Call{Query}{query, n}$ 
    \If{$pred \neq \text{NULL}$}
        \State $result \gets pred$
        \State \textbf{break} 
    \EndIf
    \State $n \gets n - 1$ 
\EndWhile

\State \Return $result$ 
\end{algorithmic}
\end{algorithm}

%% file: exps.tex
\section{Experiments}
\subsection{Implementation Details}
We selected a diverse range of models, varying in scale, architectural design, and training approaches, to ensure a thorough evaluation, including LLaMA-7B~\citep{touvron2023llama}, LLaMA-2-7B~\citep{touvron2023llama2}, ChatGLM3-6B~\citep{du2022glm}, LLaMA-2-13B, CodeLLaMA-7B~\citep{roziere2023code}, CodeLLaMA-13B, and instruction-tuned variants such as Alpaca-7B and Alpaca-CNN/DM-7B, fine-tuning details are provided in the Appendix~A.4.  
We use one RTX-3090 GPU for all 7B models, while the larger 13B models necessitate four RTX-3090 GPUs and the accelerate\footnote{https://github.com/huggingface/accelerate} library.

    
    


\subsection{Datasets \& Metrics}
To validate the effectiveness of our method in accelerating text generation for LLMs, we concentrated on two tasks: text summarization and code generation, utilizing datasets such as CNN/Daily Mail (CNN/DM)~\citep{cnndm-data}, Extreme Summarization (XSum)~\citep{xsum}, and the HumanEval~\citep{humaneval}. For additional details on the evaluation settings, please see Appendix A.5.
We employ the speed-up ratio as the evaluation metric, which is calculated by dividing the inference time of the autoregressive process by the inference time of the ANPD process, under identical conditions across all samples (For summarization tasks, we use a sample size of 1000 to ensure statistical significance, as recommended by \citep{zhang2023draft}). This metric intuitively demonstrates the performance improvement in speed when using the ANPD algorithm.

\subsection{ Main Results}
In Table~\ref{tab:t1}, we present a comparative analysis that outlines the acceleration benefits for various models and datasets. 
We have selected ~\citep{zhang2023draft} for comparison. Not only are their experimental datasets and models aligned with ours, but their methodologies are also open-sourced to facilitate easy replication.
The prompts used with these models are comprehensively documented in Appendix~A.5 to facilitate further examination and ensure the reproducibility of the results reported in this paper.

As illustrated in Table~\ref{tab:t1}, the ANPD algorithm consistently accelerates inference across various models, including the base LLM, the instruction-fine-tuned Alpaca, and the model fine-tuned with dataset-specific instructions, indicating its robustness and efficiency in accelerating text generation. Remarkably, for the LLaMA-7B model, ANPD can speed up the inference speed over 2.0$\times$, which is still valid on LLaMA2. 
Our method achieves a twofold (2.9088$\times$ vs. 1.3293$\times$) increase in acceleration compared to \citep{zhang2023draft} on the LLaMA-2-13B.
Despite the ChatGLM3 model having a significantly larger vocabulary (nearly twice that of LLaMA, the token/word ratio will be closer to 1), our ANPD algorithm still achieves a speed-up of 1.7046$\times$ and 1.6647$\times$ for CNN/DM and XSum, respectively. 
In ChatGLM3, ANPD's predictive mechanism primarily leverages the associative relationships between phrases and individual words, rather than engaging in token-level predictions within the words themselves.
So, ANPD maintains robustness and consistently enhances inference speeds across varied LLMs. 
Owing to the presence of a high occurrence of correlated patterns in code writing tasks, which significantly enhanced the prediction accuracy of the ANPD algorithm. The ANPD algorithm was able to achieve a substantial speed-up of 3.6665$\times$ on the HumanEval, but \citep{zhang2023draft} only has a speed-up of 1.6758$\times$ for CodeLLaMA-13B.

\begin{table}[h!]
    \centering
    \resizebox{\columnwidth}{!}{
    \begin{tabular}{|c|c|c|c|}
    \hline
    Model  & shot& CNN/DM & XSum\\ 
    \hline
    LLaMA-7B & 1  & 2.7455x & 3.1195x\\
    Alpaca-7B  & 0  & 2.5566x & 2.3022x \\
    Alpaca-CNN/DM-7B  &0 & 1.9481x & 2.0561x \\
    \hline
    LLaMA-2-13b~\citep{zhang2023draft} &1 & 1.3293x &1.2801x \\
    LLaMA-2-7B &1 & 2.8604x &2.7973x  \\
    LLaMA-2-13B &1 & 2.9088x &2.6063x \\
    \hline
    ChatGLM3-6B &0  & 1.7046x & 1.6647x \\
    \hline
    \hline
    Model  & shot& \multicolumn{2}{c|}{HumanEval}\\ 
    \hline
    CodeLLaMA-13B~\citep{zhang2023draft}  &0 & \multicolumn{2}{c|}{1.6758x}  \\
    CodeLLaMA-7B  &0 & \multicolumn{2}{c|}{3.5985x} \\
    CodeLLaMA-13B  &0 & \multicolumn{2}{c|}{3.6665x}  \\
    
    \hline

    \end{tabular}
    }
    \caption{The comparison of acceleration effects on different models and datasets.}
    \label{tab:t1}
\end{table}

\subsection{Ablation Study}
\begin{figure}[h]
    \centering
    \includegraphics[width=0.45\textwidth]{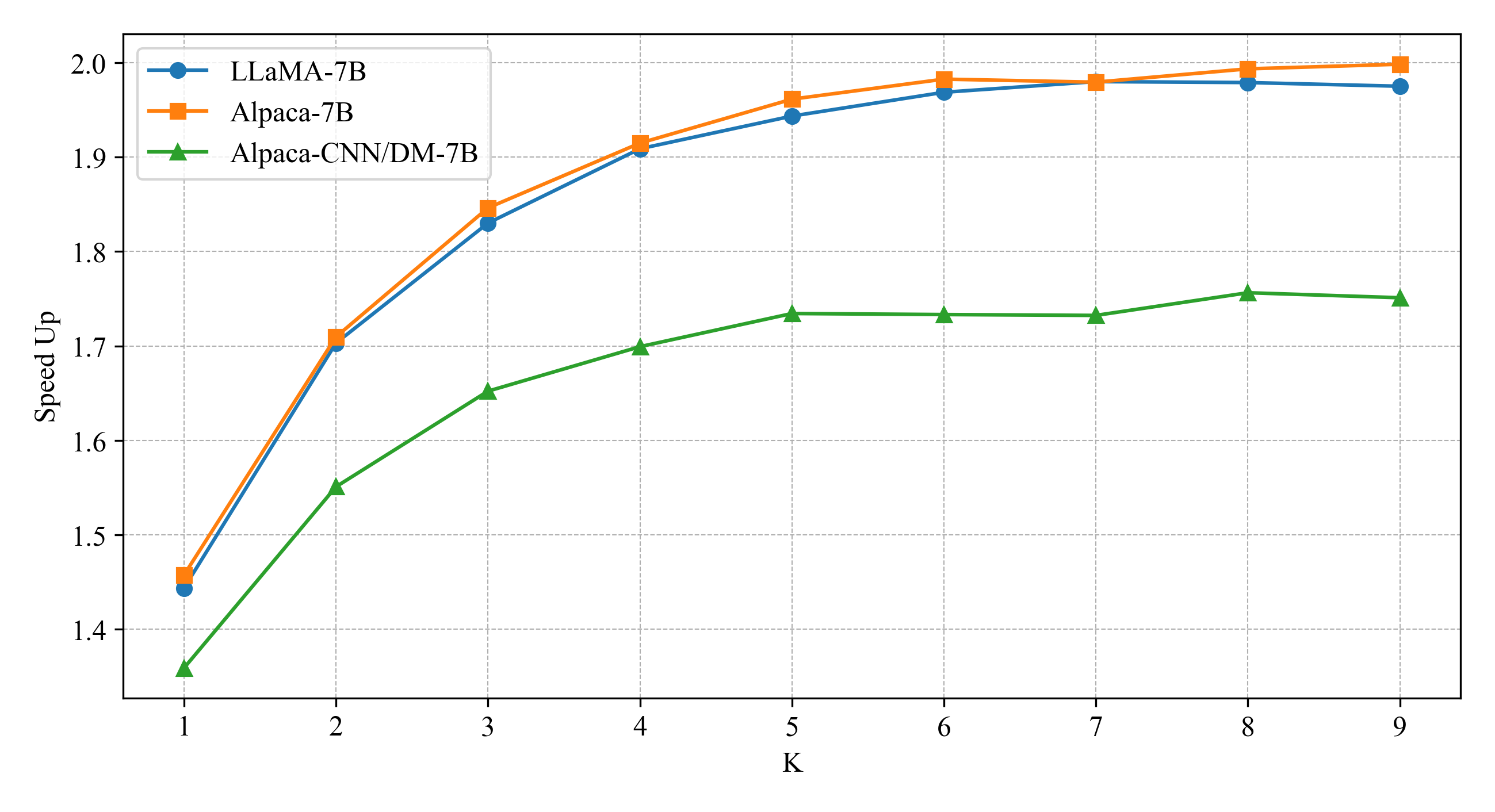}
    \caption{Speed up ratio of LLM for different $K$.}
    \label{fig:4}
\end{figure}

We conduct an analysis of hyperparameters on CNN/DM dataset, focusing primarily on \( K \) and \( N \). 
In Figure \ref{fig:4}, we set \( N \) to 2, and perform a comparative analysis of the parameter \( K \). Our findings indicate that increasing \( K \) contributes to a greater acceleration effect, however, the acceleration gains plateau when \( K \) lies within the range of 6 to 8. 

\begin{figure}[h]
    \centering
    \includegraphics[width=0.45\textwidth]{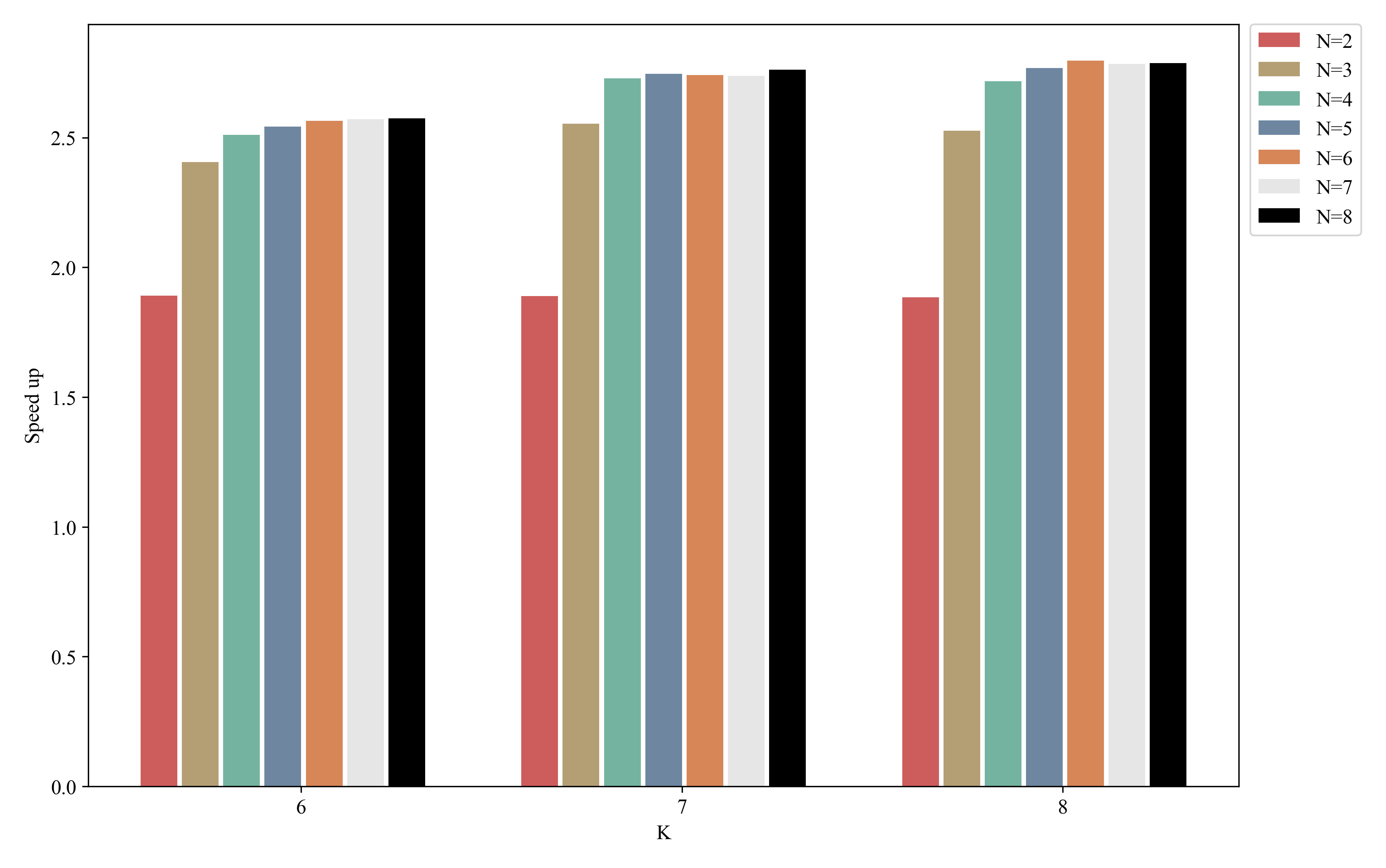}
    \caption{Decoding speed up ratio of LLaMA-7B for different $K$ and $N$.}
    \label{fig:MLN_K_exp_llama_7b}
\end{figure}

\begin{figure}[h]
    \centering
    \includegraphics[width=0.45\textwidth]{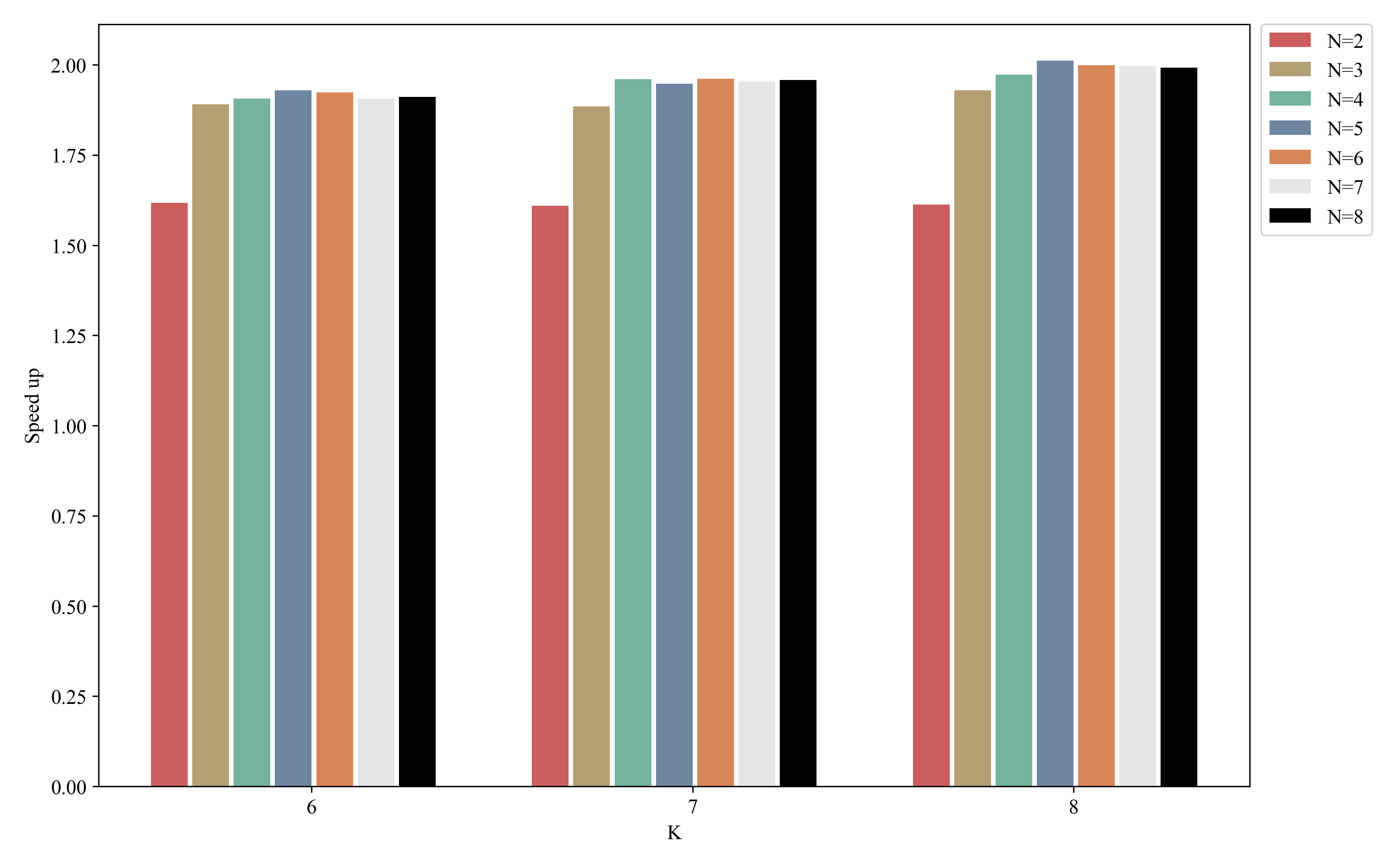}
    \caption{Decoding speed up ratio of Alpaca-CNN/DM-7B for different $K$ and $N$.}
    \label{fig:MLN_K_exp_alpaca.en.cnndm}
\end{figure}

Based on the experiment in Figure \ref{fig:4}, we selected 6, 7, and 8 for  \( K \) to conduct further hyperparameter combination experiments, as illustrated in Figures~\ref{fig:MLN_K_exp_llama_7b} and~\ref{fig:MLN_K_exp_alpaca.en.cnndm}.
The experimental results indicate that the Multi-Level N-gram (MLN) approach enhances inferential speed as the parameter \( N \) increases. However, beyond 
$N =5$, further increments in \( N \) yield no significant additional gains. Additionally, the effect of the parameter \( K \) on acceleration is relatively stable; as shown in Figure \ref{fig:4}, the acceleration effect reaches a plateau within the range of 6 to 8 for \( K \). These findings are consistent across different models with different  \( N \).

Based on the empirical evidence presented in Figure \ref{fig:MLN_K_exp_llama_7b} and Figure \ref{fig:MLN_K_exp_alpaca.en.cnndm}, a pragmatic choice for \( N \) and \( K \) can be posited at \( N = 5 \) and \( K = 7 \) respectively. 
The analogous experiments pertaining to the HumanEval dataset have been relegated to Appendix~A.6 for reference, similar conclusions can also be observed in this dataset. While employing the Multi-Level N-gram (MLN) has improved the accuracy of draft predictions, we have also carried out distinct experiments (Figure~10, Appendix~A.6)  using N-gram modules without MLN, to demonstrate that simply enlarging the value of \( N \)  is not effective.

\subsection{Case Study}
\begin{figure*}[!htbp]
    \centering
    \includegraphics[width=0.8\textwidth]{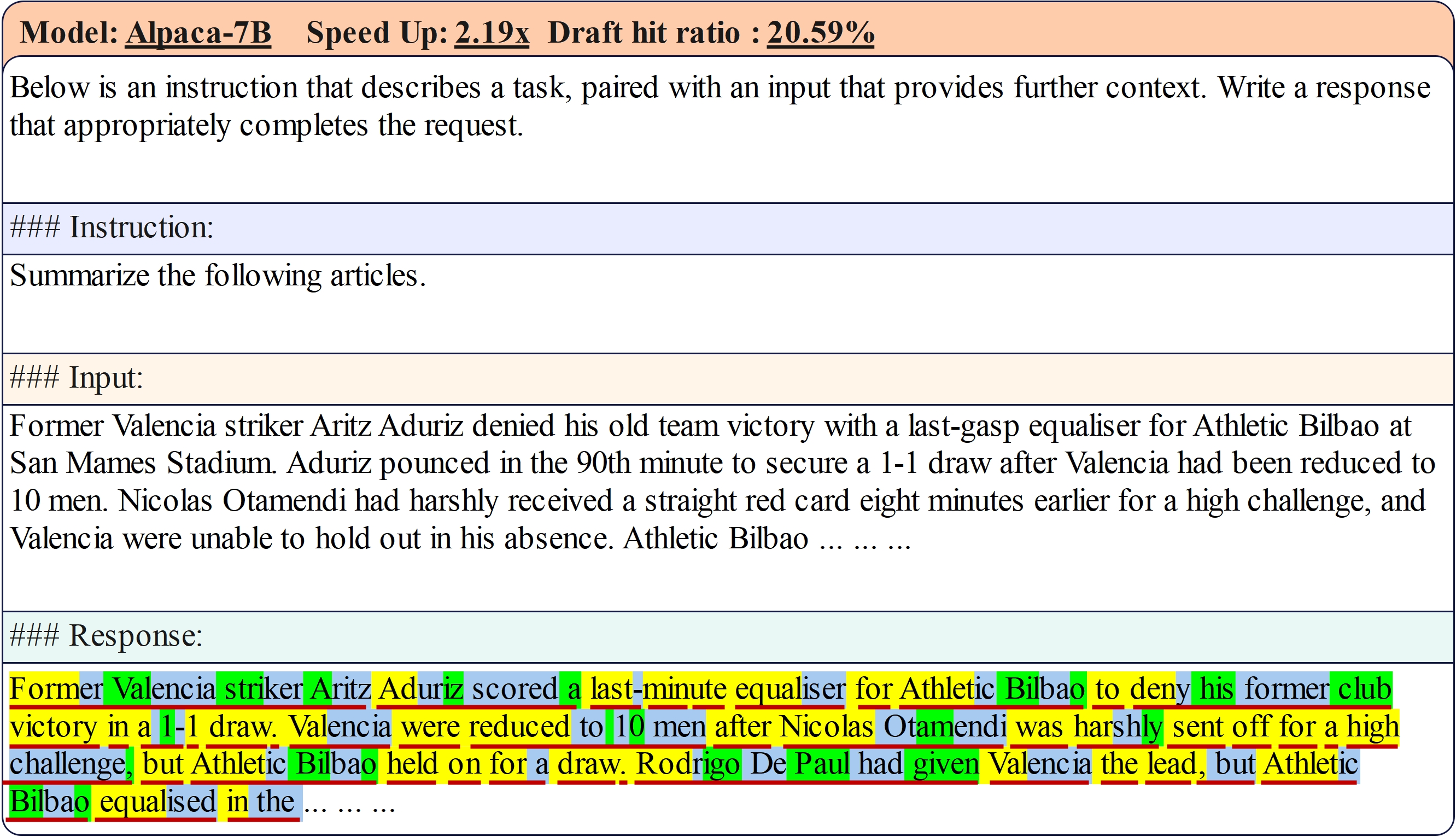}
    \caption{Visualizing the step-by-step inference process of ANPD: An example from CNN/DM.}
    \label{case:1}
\end{figure*}
Figure \ref{case:1} showcases a detailed example of the ANPD inference process, utilizing the Alpaca-7B model on a sample from the CNN/DM test set. The Alpaca-7B model, which has been fine-tuned with instructions, was chosen due to its broad applicability in practical scenarios. In this example, the ANPD algorithm is configured with $N=5$ and $K=7$, achieving a 2.19$\times$ decoding speed-up compared to the original autoregressive process, with a draft text pass rate (Draft hit ratio, $\alpha$) of 20.59\% in the LLM verification phase. Based on the hit ratio, we can derive the theoretical upper bound of acceleration as $(\alpha \times K) + 1 $, we can calculate that the theoretical speed-up is 2.44, as the loss caused by implementation problems will be slightly higher than the actual acceleration rate. The Figure \ref{case:1} uses red underlines to represent a decoding step, including drafting and verification, with the yellow background indicating the beginning of one step. Light blue and green backgrounds mark the draft content that has passed verification. This example demonstrates that inference acceleration primarily benefits from the combination of names (e.g., \textcolor{red}{\_Athlet, ic, \_Bil, ba, o}), partial words(e.g., \textcolor{red}{\_har, sh, ly}), and phrases (e.g., \textcolor{red}{\_reduced, \_to}), aligning with the motivation behind the ANPD algorithm. The ANPD can quickly capture the association between tokens and words based on this information, and establish the prediction model, thus accelerating the end-to-end decoding process.

\subsection{User Friendly}
As ANPD does not involve additional deep learning models or plug-in databases, it does not require complex initialization processes and environment configuration installations. Consequently, users can employ it directly and with great convenience, as illustrated in Listing~\ref{lst:python_code}. We plan to release the associated open-source software packages on GitHub\footnote{https://github.com/oujieww/ANPD}, making them accessible for everyone to utilize and contribute to.

\begin{lstlisting}[language=Python, caption=Python example, label=lst:python_code]
from anpd import anpd_llm
# import other libraries as usual
model = AutoModel.from_pretrain()
model = anpd_llm(model, n=5, k=7)
prompt = "Hello,World!"
result = model.gen(prompt)
\end{lstlisting}

%% file: conc.tex
\section{Conclusion}
In this paper, we presented the ANPD algorithm, a novel and lossless approach to accelerate the Large Language Models (LLMs) inference. This algorithm implements an adaptive N-gram modeling strategy, reducing the necessity for large corpora and eliminating the requirement to build an additional deep-learning draft language model. 
The Multi-Level N-gram (MLN) strategy not only enhances draft output accuracy but also further boosts efficiency.
Our empirical studies across various models and datasets validate the ANPD algorithm's effectiveness, with a remarkable peak acceleration of up to 3.67$\times$ achieved. 
The ANPD algorithm has demonstrated its potency as a powerful tool for enhancing the efficiency of LLMs.  As a plug-and-play module, it enables more extensive and pragmatic use of LLMs in various real-world contexts.

\textbf{Future Works.} We believe that ANPD can be further enhanced in two key aspects:
\begin{itemize}
    \item [1.] 
    Incorporating the specific characteristics of individual LLMs (e.g., LLaMA, ChatGLM) by creating features tailored to different LLMs to further accelerate the inference performance.
    \item [2.] 
    Exploring the possibility of generating multiple tokens in parallel during the LLMs verification process to further accelerate the inference performance.
\end{itemize}

\section{Acknowledgements}
This research is supported by the National Key Research and Development Program of China with Grant ID 2018AAA0103203 and the Chengdu Science and Technology Project with Grant ID 2022-YF05-02014-SN. This research is also supported by Huawei MindSpore Team for providing technical assistance and experience sharing.